\documentclass[letterpaper]{article} 
\usepackage{aaai25}  
\usepackage{times}  
\usepackage{helvet}  
\usepackage{courier}  
\usepackage[hyphens]{url}  
\usepackage{graphicx} 
\usepackage{natbib}  
\usepackage{caption} 
\usepackage{siunitx} 
\usepackage{caption} 
\usepackage{algorithm}
\usepackage{algorithmic}
\usepackage{xcolor}
\usepackage{amsmath}
\usepackage{amsthm}
\usepackage{tabularx} 
\usepackage{booktabs}
\usepackage{float}

\usepackage{newtxtext,newtxmath}

\newtheoremstyle{break}
  {\topsep}   
  {\topsep}   
  {\itshape}  
  {}          
  {\bfseries} 
  {:}         
  {\newline}  
  {}          
\theoremstyle{break}

\usepackage{newfloat}
\usepackage{listings}
\DeclareCaptionStyle{ruled}{labelfont=normalfont,labelsep=colon,strut=off} 
\floatstyle{ruled}
\newfloat{listing}{tb}{lst}{}
\floatname{listing}{Listing}

\title{GACO-CAD: Geometry-Augmented and Conciseness-Optimized CAD Model Generation from Single Image}

\author{
    Yinghui Wang\textsuperscript{\rm 1},
    Xinyu Zhang\textsuperscript{\rm 1}\thanks{Corresponding author: xyzhang@sei.ecnu.edu.cn.},
    Peng Du\textsuperscript{\rm 2}\thanks{Corresponding author: dp@zju.edu.cn.}
}
\affiliations{
    \textsuperscript{\rm 1}East China Normal University, China\\
    \textsuperscript{\rm 2}Zhejiang University, China
}

\usepackage{bibentry}

\begin{document}

\maketitle

\begin{abstract}


Generating editable, parametric CAD models from a single image holds great potential to lower the barriers of industrial concept design. However, current multi-modal large language models (MLLMs) still struggle with accurately inferring 3D geometry from 2D images due to limited spatial reasoning capabilities. We address this limitation by introducing GACO-CAD, a novel two-stage post-training framework. It is designed to achieve a joint objective: simultaneously improving the geometric accuracy of the generated CAD models and encouraging the use of more concise modeling procedures. First, during supervised fine-tuning, we leverage depth and surface normal maps as dense geometric priors, combining them with the RGB image to form a multi-channel input. In the context of single-view reconstruction, these priors provide complementary spatial cues that help the MLLM more reliably recover 3D geometry from 2D observations. Second, during reinforcement learning, we introduce a group length reward that, while preserving high geometric fidelity, promotes the generation of more compact and less redundant parametric modeling sequences. A simple dynamic weighting strategy is adopted to stabilize training. Experiments on the DeepCAD and Fusion360 datasets show that GACO-CAD achieves state-of-the-art performance under the same MLLM backbone, consistently outperforming existing methods in terms of code validity, geometric accuracy, and modeling conciseness.

\end{abstract}

%

\section{Introduction}
Computer-Aided Design (CAD) provides powerful support for industrial part design through a formalized language of modeling operations. In modern smart manufacturing, nearly all products are digitally defined and iterated as CAD models before entering mass production. However, mastering this operational language entails a steep learning curve: engineers typically require years of training to fluently manipulate parameters, constraints, and feature trees to translate abstract geometric concepts into manufacturable and maintainable digital prototypes. This difficulty poses a critical bottleneck within the established "design–verify–manufacture" cycle.

Consequently, the ability to rapidly generate accurate and editable CAD models—much like text or image generation—has emerged as a key challenge for both industry and academic research. Traditional 3D generation methods~\cite{sun2023dreamcraft,xiang2025structured3d,ling2025scenethesis,cai2020learninggradient} mostly produce non-procedural geometric representations, lacking parametric modeling histories and thus unsuitable for downstream industrial editing. A number of recent studies~\cite{xu2024CADMLLM,Yuan202OpenECAD,Wang2025cadgpt,kolodiazhnyi2025cadrille,rukhovich2024cadrecode,li2025seekcad,Kerbl2023GaussianSplatting} have adopted the Large Language Models (LLMs) framework to treat CAD generation as a code-generation problem. Despite this progress, the generated outputs currently lack the necessary geometric precision and code correctness required for real-world industrial applications.

The core difficulty stems from the need to infer accurate 3D spatial relationships using just a single 2D image. Existing Multi-modal Large Language Models (MLLMs) are predominantly trained on tasks like image-text matching, visual question answering and document understanding, which do not explicitly encode the 3D geometry implied by the images. Consequently, these models exhibit limited spatial reasoning abilities for single-view reconstruction. Although some studies ~\cite{wu2025spatialmllm,cheng2024spatialrgpt,chen2024spatialvlm} have explored spatial reasoning in general vision-language contexts, such efforts remain limited in CAD model generation.

Meanwhile, reinforcement learning (RL) as a key post-training technique for large language models (LLMs) ~\cite{yu2025dapo,zheng2025gspo,shao2024grpo,rafailov2024dpo} has recently shown strong potential in CAD generation ~\cite{kolodiazhnyi2025cadrille} and effectively improves output quality. However, existing approaches focus exclusively on geometric accuracy and code validity while neglecting modeling conciseness—a critical factor in industrial practice, where redundant operations reduce code readability and increase editing costs.

\begin{figure*}[t]
  \centering
  \includegraphics[width=\linewidth]{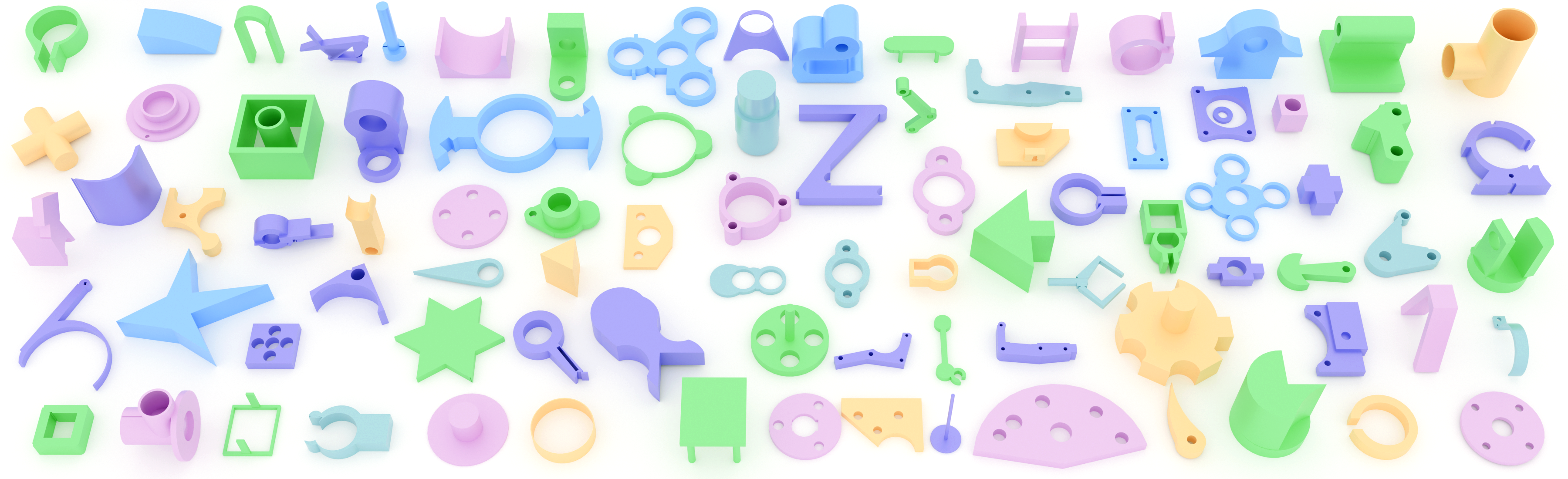}
  \caption{Demenstration of various CAD models generated by GACO-CAD.}
  \label{fig:main show}
\end{figure*}

To address the two key challenges of limited geometric understanding and redundant generation sequences, we propose GACO-CAD, a two-stage post-training framework for single-view image-to-CAD generation. In the supervised fine-tuning (SFT) stage, we introduce RGB images, depth maps, and surface normal maps as multi-channel inputs, leveraging dense geometric priors to enhance the MLLM’s 3D spatial reasoning. In the RL stage, we propose a group length reward mechanism that jointly optimizes geometric accuracy and sequence conciseness, stabilized by a simple dynamic weighting strategy. Experimental results show that GACO-CAD outperforms existing methods on both DeepCAD and Fusion360 datasets, achieving state-of-the-art performance across code validity, geometric accuracy, and modeling conciseness. Some successful CAD models generated by GACO-CAD are shown in Figure \ref{fig:main show}.The key contributions of our study are summarized as follows:

\begin{itemize}
    \item We analyze the significance of dense geometric priors in enhancing space understanding for MLLM-based single-view CAD generation and introduce these priors during MLLM training and inferring. By integrating these priors during both MLLM training and inference, we effectively improve the geometric accuracy of the generated CAD models.

    \item We design a group length reward mechanism that explicitly represents generation conciseness as an optimization objective, expanding a new evaluation dimension for generating CAD models.

    \item Our approach outperforms previous MLLM-based methods on both Deepcad and Fusion360 datasets under the same MLLM backbone, substantially improving the performance of single-view CAD generation task.

    
    
    
\end{itemize}

\section{Related Work}

\noindent \textbf{CAD Model Generation.}
Editable CAD models are typically represented as parametric operation sequences or code with modeling APIs~\cite{au2024cadquery}. Early works~\cite{Wu2021DeepCAD,xu2022skexgen,ganin2021cadaslan} employed Transformer architectures~\cite{vaswani2023attention} to autoregressively generate modeling command sequences, laying the foundation for procedural CAD generation. Recent studies have further introduced hierarchical structures ~\cite{xu2023hierarchical} or diffusion-based mechanisms~\cite{Ma2024DrawStep,chen2025cadcrafter,Liu2025HoLa} to generate CAD sequences conditioned on text or images. However, these approaches often rely on non-intuitive, low-level command representations, limiting code readability and editability.

With the rise of LLMs and MLLMs, researchers have begun framing CAD generation as a structured code generation task. Some works~\cite{ocker2025ideacad,li2025seekcad} directly leverage the reasoning capabilities of general-purpose large models via agent-based systems that decompose complex design requirements. However, due to the lack of task-specific training, they suffer from low Pass@1 accuracy and require extensive prompt engineering and multi-turn interactions, leading to high inference costs. Another line of research focuses on dedicated training: OpenECAD~\cite{Yuan202OpenECAD} firstly explore image-to-CAD generation on multimodal large models, though it was limited by early model capabilities and context length. Subsequent efforts~\cite{xu2024CADMLLM,rukhovich2024cadrecode,2024text2cad} have expanded input modalities and architectural adaptations, but still struggle to meet industrial demands for precision and robustness. 

\noindent \textbf{Spatial Understanding in MLLMs.}
Current MLLMs~\cite{wang2024qwen2vl, zhu2025internvl3,bai2025qwen25vl} excel on standard image-text understanding benchmarks, but exhibit clear limitations in spatial perception: they struggle to accurately infer 3D positions, distances, or geometric structures from a single 2D image. This stems primarily from their pre-training objectives, which emphasize 2D image-text alignment rather than geometrically precise spatial understanding.

To bridge this gap, some works~\cite{wu2025spatialmllm,cheng2024spatialrgpt,Liu2025SSR,chen2024spatialvlm} have attempted to inject depth estimation or spatial relation data during pre-training or fine-tuning to endow MLLMs with better 3D reasoning capabilities. In the CAD domain, CAD-GPT~\cite{Wang2025cadgpt} introduced an innovative spatial localization mechanism using dedicated tokens to enhance contextual awareness, but this approach relies on known modeling histories and is unsuitable for open-ended generation scenarios. Recent advances in dense geometric estimation ~\cite{he2024lotus,hu2024metric3dv2,2024depth_anything_v2,wang2025moge2} have made it possible to reliably estimate high-fidelity depth and surface normal maps from a single RGB image. Building on this foundation, we incorporate depth and normal maps as geometric priors into the MLLM input to enhance the model’s shape perception and generalization.

\noindent \textbf{Reinforcement Learning for LLM Post-Training.}
Reinforcement learning plays a pivotal role in post-training large models, widely applied to human preference alignment, tool usage, and code generation. Early approaches like RLHF-PPO~\cite{schulman2017ppo} require maintaining multiple networks and are complex to implement. Later methods such as DPO~\cite{rafailov2024dpo} reformulate preference data into a cross-entropy loss, simplifying training but lacking support for online exploration. Algorithms like GRPO~\cite{shao2024grpo}, DAPO~\cite{yu2025dapo}, and CPPO~\cite{lin2025cppo} improve efficiency through group sampling and self-normalized advantage estimation for on-policy learning. The recent GSPO~\cite{zheng2025gspo} further introduces sequence-level importance sampling and clipping, achieving enhanced training stability without sacrificing simplicity.

Despite these advances in general domains, RL applications in CAD generation remain in the early stage of exploration. Cadrille~\cite{kolodiazhnyi2025cadrille} was the first to apply RL training to MLLM-based CAD generation, using rewards based on code validity and IoU to improve output quality. We argue that RL’s potential in this domain is far from fully exploited. Building upon these foundation, we are the first to explicitly optimize for modeling conciseness in the RL stage, designing a holistic reward function that not only enforces geometric correctness but also encourages the generation of more concise and efficient modeling programs.
\section{Preliminary}

\subsection{LLM Supervised Fine-Tuning}
Supervised Fine-Tuning (SFT) adapts a pretrained model to a target output distribution using labeled data, thereby improving downstream accuracy and consistency. It maximizes the likelihood of the target sequence given the context. Let $x$ denote the multimodal context and $y=(y_1,\dots,y_T)$ the desired cadquery code~\cite{au2024cadquery}. The negative log-likelihood loss is
\begin{equation}
\mathcal{L}_{\text{SFT}}(\theta)=-\mathbb{E}_{(x,y)\sim\mathcal{D}}\sum_{t=1}^{T}\log\pi_{\theta}(y_t\mid y_{<t},x),
\end{equation}
where $\mathcal{D}$ is the supervised dataset, $\theta$ the model parameters, and $\pi_{\theta}$ the token-level probability. In this paper, $x$ contains a single RGB image, depth and normal maps, plus brief text.
\subsection{LLM Reinforcement Learning}
After SFT, Reinforcement Learning refines the policy without additional labels by optimizing a quality-oriented reward. We adopt Group Sequence Policy Optimization (GSPO)\cite{zheng2025gspo}, which performs group-wise updates using sequence-level rewards. The GSPO objective $\mathcal{J}_{\text{GSPO}}(\theta)$ is
\begin{equation}
\mathbb{E}\Bigl[\,\frac{1}{G}\sum_{i=1}^{G}\min\!\bigl(s_i(\theta)\hat{A}_i,\,\mathrm{clip}(s_i(\theta),1\!-\!\varepsilon,1\!+\!\varepsilon)\hat{A}_i\bigr)\Bigr],
\end{equation}
where the expectation is taken over $x\sim\mathcal{D}$ and $G$ candidates $y_i\sim\pi_{\theta_{\text{old}}}$, $s_i(\theta)$ is the importance weight, and $\varepsilon$ the clipping radius. The advantage $\hat{A}_i$ is computed via group normalization
\begin{equation}
\hat{A}i = \frac{r(x, y_i) - \text{mean}\bigl({r(x, y_i)}_{i=1}^G\bigr)}{\text{std}\bigl({r(x, y_i)}_{i=1}^G\bigr)},
\end{equation}
with $r(\cdot)$ a rule-based reward that evaluates the quality of the generated cadquery code.
\section{Method}
\begin{figure*}[t]     
  \centering
  \includegraphics[width=\linewidth]{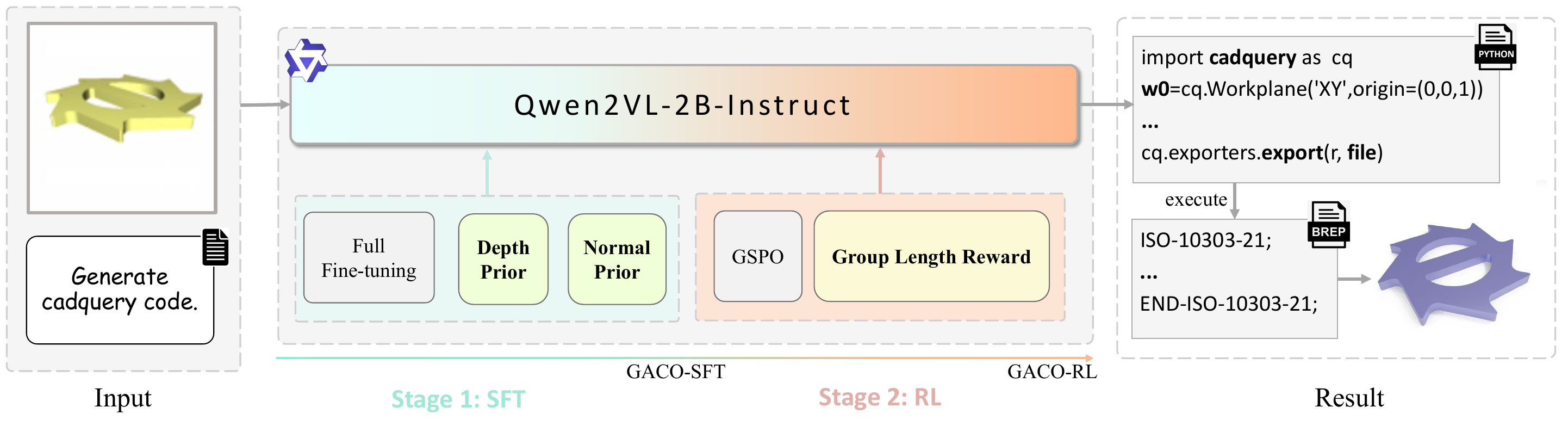}
  \caption{\textbf{Main Pipeline of Two-stage Training.} Starting from pre-trained weights, we obtain the final model through two training stages: SFT and RL. In the SFT stage, we perform full fine-tuning with depth and normal priors; in the RL stage, we optimize with the GSPO algorithm incorporating a length-reward rule. After training, the model takes a single-view image and simple text as input and generates target Python code, whose execution yields the final B-rep file.}
  \label{fig:main pipeline}
\end{figure*}

As illustrated in Figure~\ref{fig:main pipeline}, our training proceeds in two tightly-coupled stages.
First, SFT injects geometric priors—depth and surface normals—into a pre-trained MLLM, yielding an initial policy (GACO-SFT) that already produces executable CAD code from a single image and brief prompt.
Second, RL refines this policy with a novel group length reward that trades off geometric accuracy against code conciseness, producing the final model (GACO-RL) capable of generating compact and accurate Python scripts whose execution returns the target B-rep.

\subsection{Enhancing Spatial Understanding with Geometric Priors}

In this subsection, we propose a geometric prior-based approach to enhance the spatial understanding of MLLM for single-view CAD model generation. We first analyze the non-uniqueness of 3D reconstruction from a single 2D image in \S\,\ref{sec:method-depth}, emphasizing the importance of depth priors. Then, in \S\,\ref{sec:sft}, we detail how geometric priors are integrated into the SFT stage.

\subsubsection{Depth Prior in Single-View 3D Reconstruction}
\label{sec:method-depth}

\begin{figure*}[t]     
  \centering
  \includegraphics[width=\linewidth]{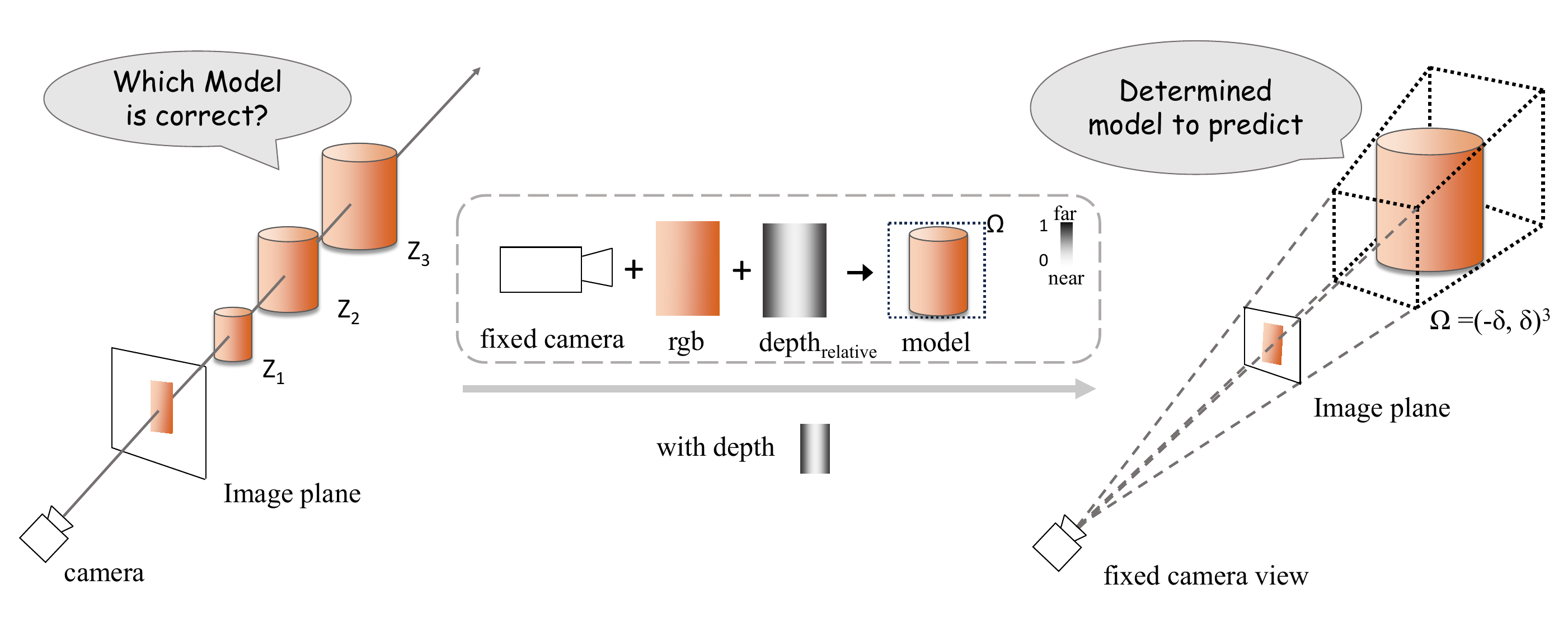}
  \caption{\textbf{The Role of Deep Prior in Monocular 3D Reconstruction.} When the camera parameters and the limited output model bounding box are defined, the introduction of the relative depth map theoretically provides sufficient conditions for accurately reconstructing the 3D model of the visible part in a single view.}
  \label{fig:pixel2world}
\end{figure*}

Reconstructing accurate 3D structures from a single RGB image is an ill-posed problem due to its geometric non-uniqueness. For a pixel coordinate $\mathbf{x} = (u, v)$, its corresponding 3D point $\mathbf{X}_w$ satisfies:
\begin{equation}
\mathbf{X}_w = \mathbf{R}^{-1}\left( Z \mathbf{K}^{-1} \tilde{\mathbf{x}} - \mathbf{t} \right), \quad Z > 0,
\end{equation}
where $\tilde{\mathbf{x}} = (u, v, 1)^T$ is the homogeneous coordinate, $\mathbf{K}$ is the camera intrinsic matrix, and $(\mathbf{R}, \mathbf{t})$ denotes the extrinsic parameters. Since both depth $Z$ and camera parameters are unknown, there exist infinitely many possible 3D points corresponding to a single pixel, making single-view 3D reconstruction inherently ambiguous.

In practice, training data for image-to-CAD generation is often rendered using fixed camera intrinsics and extrinsics, and the ground-truth 3D models are constrained within a bounding box $\Omega = [-\delta, \delta]^3$, as illustrated in Figure \ref{fig:pixel2world}. Under this Condition, the depth variable $Z$ is implicitly restricted to a known interval. If the relative depth ordering of surface points is provided, the 3D coordinates of visible regions in the single view can be uniquely determined. The depth map $Z_{\text{rel}}(u, v)$ offers this relative ordering, serving as a key geometric constraint.

It is crucial to highlight that MLLMs do not engage in explicit back-projection. Instead, their utilization of depth information is grounded not in geometric reasoning but in the alignment of RGB-depth statistical co-occurrences throughout the comprehensive pre-training and fine-tuning phases. Despite this, depth information remains vital in conveying geometric relationships. By transforming "geometric correctness" into an easily learnable mapping characterized by "high feature co-occurrence probability," the complexity of reconstructing the 3D structure from solely RGB views is diminished. Guided by this understanding, we incorporate depth priors as additional input during training to augment the model's spatial comprehension and enhance the accuracy of single-view CAD generation.

\subsubsection{Supervised Fine-Tuning with Geometric Priors}
\label{sec:sft}

\begin{figure}[t]    
  \centering
  \includegraphics[width=\linewidth]{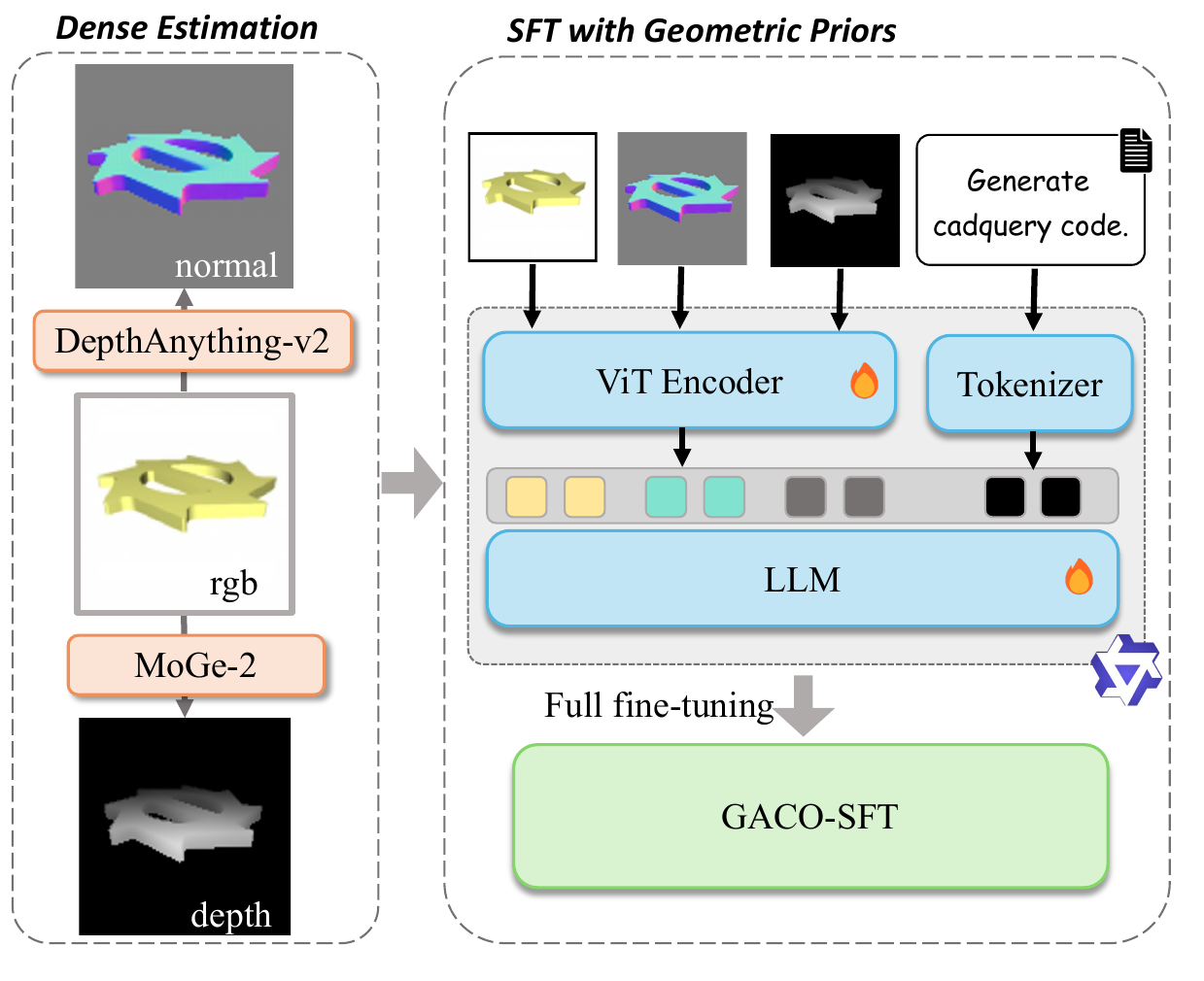}  
  \caption{\textbf{SFT Pipeline.} We use Moge-2 and DepthAnything v2 to estimate normal map and depth map. All three images are processed by the shared ViT encoder, concatenated with the text, and fed into the LLM and get the first-stage model: GACO-SFT.}
  \label{fig:sft Pipeline}
\end{figure}

During SFT, in addition to the RGB image $I^{\text{rgb}}$ and depth map $I^{\text{dep}}$, we introduce surface normals $I^{\text{nor}}$ as an extra geometric modality. While depth provides relative ordering among pixels, normals offer texture-independent surface orientation cues, helping the model focus on shape boundaries and curvature changes without being distracted by RGB texture ambiguities.

All three visual modalities $(I^{\text{rgb}}, I^{\text{dep}}, I^{\text{nor}})$ are processed by the shared pretrained vision encoder $E_v(\cdot)$ to extract patch-level features: $H^{rgb} = E_v(I^{rgb}), H^{\text{dep}} = E_v(I^{\text{dep}}), H^{\text{nor}} = E_v(I^{\text{nor}})$. These features are then projected into the LLM embedding space via the shared projection layer $P(\cdot)$: $Z^{rgb} = P(H^{rgb}), Z^{\text{dep}} = P(H^{\text{dep}}), Z^{\text{nor}} = P(H^{\text{nor}})$. The final input sequence is a concatenation of visual tokens and text prompt embeddings: $x = [Z^{rgb}, Z^{\text{dep}}, Z^{\text{nor}}, T]$, where $T$ denotes the brief text prompt embedding, and $[\cdot]$ represents concatenation along the token dimension.

The training pipeline is illustrated in Figure \ref{fig:sft Pipeline}. Surface normals are estimated using MoGe-2~\cite{wang2025moge2}, and depth maps are generated using DepthAnything v2~\cite{2024depth_anything_v2}.

\subsection{Accuracy–Simplicity Balanced Reinforcement Learning}

\label{sec:rl}
\begin{figure}[t]       
  \centering
  \includegraphics[width=\linewidth]{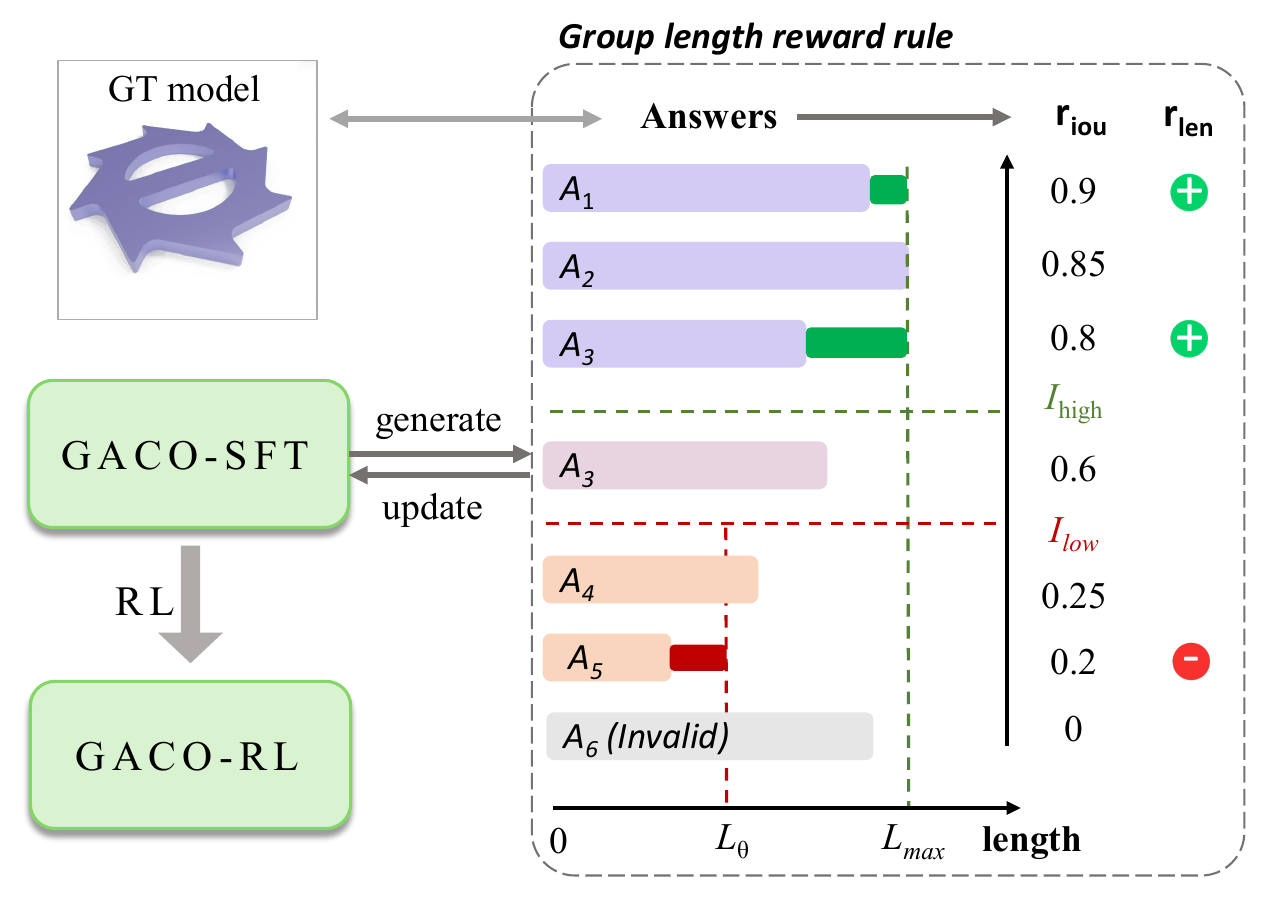}  
  \caption{\textbf{Reinforcement Learning with Group Length Reward Rule.} We conducted RL training in GACO-SFT. In the generated responses, those with high IoU accuracy would be rewarded based on the group relative token length, while responses with low IoU and below the length threshold would be penalized. Finally, we obtained the two-stage model: GACO-RL.}
  \label{fig:len reward}
\end{figure}

While supervised fine-tuning aligns the model well with task objectives, it struggles to optimize for conciseness. Redundant operations reduce code interpretability and increase editing costs, while long sequences incur higher computational and memory overhead. Without expert-annotated concise data, optimizing for simplicity is challenging under pure supervision.

Inspired by GSPO~\cite{zheng2025gspo}, we propose a group relative length reward mechanism that balances geometric accuracy and code simplicity via reinforcement learning.

\subsubsection{Group Length Reward Function}

We define a rule-based reward function $r(x, y)$ composed of three terms: code validity, geometric accuracy, and sequence length:
\begin{equation}
r(x, y) = \lambda_{\text{len}} r_{\text{len}}(x, y) + \lambda_{\text{iou}} r_{\text{iou}}(x, y) + \lambda_{\text{val}} r_{\text{val}}(x, y),
\end{equation}
where $\lambda_{\text{len}}, \lambda_{\text{iou}}, \lambda_{\text{val}}$ are corresponding weights.

Code Validity Reward $r_{\text{val}}$ checks whether the generated code is syntactically correct and executable.
Geometric Accuracy Reward $r_{\text{iou}}$ measures the IoU between the generated and ground-truth CAD models.
Length Reward $r_{\text{len}}$ encourages conciseness among high-accuracy candidates.

As shown in Figure \ref{fig:len reward}, given input $x$, the old policy generates $G$ candidate sequences $\{y_i\}_{i=1}^G$ with token lengths $\{L_i\}_{i=1}^G$ and IoU scores $\{I_i\}_{i=1}^G$. Among high-accuracy candidates ($I_i \geq I_{\text{high}}$), we define a relative length ratio:
\begin{equation}
\alpha_i = \frac{L_i}{\max_j L_j}, \quad \alpha_i \in (0, 1].
\end{equation}
The length reward is then:
\begin{equation}
r_{\text{len}}(x, y_i) = 0.5 + 0.5(1 - \alpha_i) \cdot I_i, \quad \text{if } I_i \geq I_{\text{high}}.
\end{equation}
Among accurate candidates, shorter sequences receive higher rewards.

For low-accuracy candidates ($I_i \leq I_{\text{low}}$), if the length is below a threshold $L_\theta$, a penalty is applied:
\begin{equation}
r_{\text{len}}(x, y_i) = 0.5 - 0.5 \cdot \left(1 - \frac{L_i}{L_\theta}\right), \quad \text{if } L_i < L_\theta,
\end{equation}
otherwise, a neutral reward $r_{\text{len}} = 0.5$ is given. No length guidance is applied for medium-accuracy candidates ($I_{\text{low}} < I_i < I_{\text{high}}$), ensuring the model prioritizes geometric correctness.

\subsubsection{Dynamic Weight Scheduling}
\label{sec:dynamic_weight}

The importance of each reward component varies throughout the training process. In the early stages, the model focuses on ensuring code validity and geometric accuracy, whereas later stages emphasize conciseness. To achieve a smooth transition between these objectives, we adopt a linear weight scheduling strategy:
\begin{equation}
\lambda_k(t) = \lambda_k^{\text{start}} + \frac{t}{T} \left(\lambda_k^{\text{end}} - \lambda_k^{\text{start}}\right),
\end{equation}
where $t$ denotes the current training step, $T$ is the total number of steps, and $\lambda_k^{\text{start}}, \lambda_k^{\text{end}}$ are initial and final weights for each component.

Specifically, we set $\lambda_{\text{len}}^{\text{start}} <\lambda_{\text{len}}^{\text{end}}$, $\lambda_{\text{iou}}^{\text{start}}> \lambda_{\text{iou}}^{\text{end}}$ and $\lambda_{\text{val}}^{\text{start}}> \lambda_{\text{val}}^{\text{end}}$
ensuring that early training focuses on correctness, while later stages gradually emphasize conciseness. This dynamic scheduling encourages the model to progressively balance geometric accuracy and brevity, resulting in compact yet precise CAD modeling code. 
\section{Experiments}
To validate the effectiveness of GACO-CAD, we conducted a two-stage post-training process as previously described and compared our method with existing works on the single-view CAD model generation task. In the ablation study, we further investigated the impact of two types of geometric priors and the length reward mechanism on model performance. Additionally, we analyzed the network architecture used for processing geometric prior images to demonstrate the effectiveness and design rationality of the simple integration strategy adopted in GACO-CAD.

\subsection{Experimental Setups}

\textbf{Datasets.} 
During the SFT stage, we used approximately one million samples from the Recode dataset~\cite{rukhovich2024cadrecode} as training data. All CAD models in the Recode dataset were constructed via procedurally generated CAD code within a bounded domain $\Omega = [-100, 100]^3$. Although the model distribution differs from real industrial parts, its highly randomized spatial structures provide strong training signals for MLLMs to understand spatial mappings.

In the RL stage, we utilized approximately 50k real industrial CAD models from DeepCAD~\cite{park2019deepsdf} and 3k from Fusion360~\cite{willis2021fusion360} to guide the model toward generating CAD structures that better align with real-world engineering requirements. The test data includes 8,046 samples from DeepCAD and 1,725 from Fusion360 for a comprehensive evaluation of model performance.

\textbf{Details.} 
For a fair comparison with prior work~\cite{kolodiazhnyi2025cadrille}, we adopted the same pretrained model, Qwen2VL-2B-Instruct~\cite{wang2024qwen2vl}, as the base architecture and performed two-stage post-training. The SFT stage was conducted on a single H800 GPU using the AdamW optimizer~\cite{loshchilov2019adamw} with a learning rate of $2\times10^{-4}$, weight decay of 0.01, a total batch size of 32, and 4 epochs. The RL stage was carried out on four H800 GPUs with VeRL~\cite{sheng2024verlhybridflow} framework, with a learning rate of $1\times10^{-6}$, a mini-batch size of 128, policy update every 4 iterations, and 2 epochs in total.

In RL training, we set the IoU upper threshold $I_{\text{high}}=0.8$ and lower threshold $I_{\text{low}}=0.4$. The length threshold $L_{theta}$ was set to 110, determined by the statistical mean of generated code lengths in the RL training dataset. Reward weights were dynamically adjusted during training: the length reward weight increased from $\lambda_{\text{len}}^{\text{start}}=0$ to $\lambda_{\text{len}}^{\text{end}}=0.4$; the IoU reward weight decayed from $\lambda_{\text{iou}}^{\text{start}}=0.8$ to $\lambda_{\text{iou}}^{\text{end}}=0.5$; and the validity reward weight decayed from $\lambda_{\text{val}}^{\text{start}}=0.2$ to $\lambda_{\text{val}}^{\text{end}}=0.1$. 

All input images were uniformly resized to 134×134. RGB images were rendered following the same pipeline as in cadrille~\cite{kolodiazhnyi2025cadrille} that rendering the normalized model within the bounding box $[-1,1]^3$ from a fixed camera viewpoint.

\textbf{Evaluation Metrics.}
We adopted the evaluation metrics from prior works~\cite{rukhovich2024cadrecode,kolodiazhnyi2025cadrille} to assess the quality of generated CAD code from three perspectives: Invalid Rate (IR), Intersection over Union (IoU), and Chamfer Distance (CD). IR reflects the proportion of generated code that fails to execute successfully. IoU and CD measure the geometric accuracy of the generated models. When computing CD, all models are normalized to the $[-0.5,0.5]^3$ space, and results are reported as the median CD multiplied by $10^3$.

Additionally, to evaluate the conciseness of the generated code, we introduced the Average Token Length (ATL) metric, representing the token length of the generated code text. This metric is computed using the tokenizer of Qwen2VL-2B-Instruct and is only compared among baseline methods using the same cadquery code.

\subsection{Performance}
\textbf{Quantitative Results.}
\begin{table}[t]
  \centering
  \caption{\textbf{Quantitative evaluation of single view CAD generation On DeepCAD dataset and Fusion360 Dataset.} The best results are \textbf{bold}. CD uses the median val multiplied by $10^{3}$; ATL is the average token length produced by the Qwen2VL-2B-Instruct tokenizer. $\downarrow$: lower is better; $\uparrow$: higher is better.}
  \label{tab:maincompare}
  \begin{tabular}{lcccc}
    \toprule
    Method & {IR (\%)$\downarrow$} & {IoU (\%)$\uparrow$} & {CD$\downarrow$} & {ATL$\downarrow$} \\
    \midrule
    \multicolumn{5}{c}{\textbf{DeepCAD dataset}} \\
    \midrule
    CAD-GPT                               &      &       & 9.77 &  \\
    CAD-MLLM                              &      &       & 3.77 &  \\
    Img2CAD                               &      &       & 1.61 &  \\
    CADCrafter                           &      &       & 0.82 &  \\
    cadrille-SFT(80k)   & 1.75 & 73.55 & 0.37 & 99.8 \\
    Ours-SFT(80k)       & 1.57 & 75.44 & 0.33 & 98.8 \\
    cadrille-SFT(1M)    & 1.24 & 84.54 & 0.22 & 97.6 \\
    Ours-SFT(1M)        & 1.12 & 85.67 & 0.19 & 99.2 \\
    cadrille-RL         & 0.71 & 85.99 & 0.19 & 97.7 \\
    Ours-RL             & \textbf{0.67} & \textbf{86.49} &\textbf{0.18} & \textbf{93.1} \\
    \midrule
    \multicolumn{5}{c}{\textbf{Fusion360 dataset}} \\
    \midrule
    cadrille-SFT(80k)   & 3.45 & 64.58 & 0.54 & 121.8 \\
    Ours-SFT(80k)       & 3.30 & 65.87 & 0.51 & 122.6 \\
    cadrille-SFT(1M)    & 1.57 & 77.15 & 0.22 & 126.2 \\
    Ours-SFT(1M)        & 1.39 & 78.60 & 0.20 & 125.5 \\
    cadrille-RL         & 1.20 & 78.72 & 0.20 & 125.9 \\
    Ours-RL             & \textbf{1.16} & \textbf{79.47} & \textbf{0.19} & \textbf{118.4} \\
    \bottomrule
  \end{tabular}
\end{table}
Table \ref{tab:maincompare} presents the quantitative comparison between GACO-CAD and existing methods on the single-view CAD generation task. We report the performance of three model variants: Ours-SFT(80k), a small-scale SFT model trained on ~80k Recode samples to assess the role of geometric priors under limited data; Ours-SFT(1M), the SFT model trained on the full 1M data; and Ours-RL, the final version further fine-tuned with RL. As the primary baseline, we trained three corresponding versions of the cadrille method on the same data scales for fair comparison.

Results show that our method outperforms the previous best approach in terms of IR, IoU, and CD under both 80k and 1M data scales, validating the effectiveness of introducing geometric priors for MLLM in understanding single-view geometry. Notably, the performance gain from geometric priors is more significant under small-scale training (80k). We conjecture that large-scale SFT allows the model to implicitly learn accurate depth and spatial relationships from RGB views, partially reducing the marginal benefit of geometric priors. This outcome is consistent with our mechanism analysis.

Furthermore, the RL-trained model not only improves geometric accuracy and code validity but also demonstrates advantages in ATL. On the DeepCAD and Fusion360 test sets, the token lengths of generated code are reduced by approximately 6.1\% and 5.7\%, respectively. Despite cadquery’s inherent conciseness, the proposed length reward mechanism successfully guides the policy network to generate more compact modeling code. Overall, GACO-CAD achieves state-of-the-art performance on the single-view CAD generation task across all metrics.

\textbf{Qualitative results.}
\begin{figure*}[t]
  \centering
  \includegraphics[width=\linewidth]{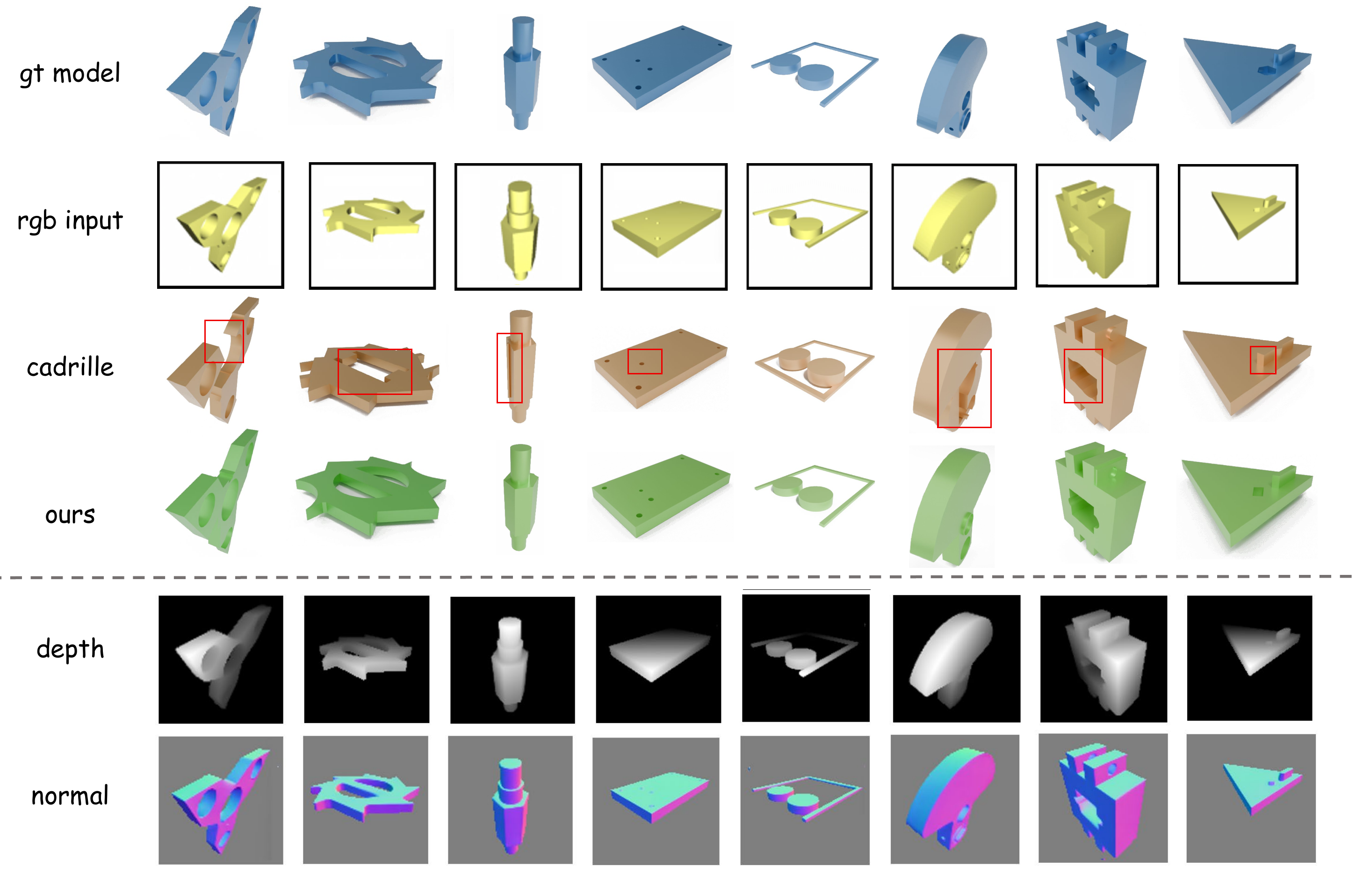}
  \caption{\textbf{Qualitative comparison of the generated CAD models}. Depth and normal inputs are only used in our method. The resolution of all input images is 134×134. Our method excels in terms of geometric details. }
  \label{fig:qualitative comparison}
\end{figure*}
Qualitative results are shown in Figure \ref{fig:qualitative comparison}. We compare the models produced by the previous best baseline and our method on the single-view CAD generation task. Owing to the injected geometric priors, our approach yields CAD models with more accurate geometry and finer, more plausible details. Although we used 134×134 ultra-low resolution images, after introducing geometric priors, MLLM can more easily distinguish key geometric information such as holes in the model.

\subsection{Ablation Study}
\label{sec:ablation}
\begin{table}[t]
  \centering
  \caption{\textbf{Ablative evaluation of geometric prior combinations On Deepcad dataset.} $\text{Ours}_{\text{wo-dn}}$: RGB only; $\text{Ours}_{\text{wo-n}}$: RGB + depth; Ours: RGB + depth + normal.}
  \label{tab:prior types ablation}
  \begin{tabular}{lccc}
    \toprule
    Method & {IR (\%)$\downarrow$} & {IoU (\%)$\uparrow$} & {CD$\downarrow$} \\
    \midrule
    $\text{Ours}_{\text{wo-dn}}$        & 1.75 & 73.55 & 0.37 \\
    $\text{Ours}_{\text{wo-n}}$         & 1.61 & 74.37 & 0.35 \\
    $\text{Ours}$                       & 1.57 & 75.44 & 0.33 \\
    \bottomrule
  \end{tabular}
\end{table}
\textbf{Types of Geometric Priors.}
To analyze the contribution of the two types of geometric priors (depth and normal maps), we trained three model variants on 80k Recode data and conducted ablation experiments on the DeepCAD test set. As shown in Table \ref{tab:prior types ablation}, the model using only RGB input ($\text{Ours}_{\text{wo-dn}}$) performs the worst; introducing depth priors ($\text{Ours}_{\text{wo-n}}$) brings some improvement; and the full model using both depth and normal priors performs the best, validating the complementarity of multi-modal geometric information.

\begin{table}[t]
  \centering
  \caption{\textbf{Ablative evaluation of length reward rule On Deepcad dataset.} $\text{Ours}_{\text{wo-len}}$: RL without length reward.}
  \label{tab:length ablation}
  \begin{tabular}{lcccc}
    \toprule
    Method & {IR (\%)$\downarrow$} & {IoU (\%)$\uparrow$} & {CD$\downarrow$} & {ATL$\downarrow$} \\
    \midrule
    $\text{Ours}_{\text{wo-len}}$        & 0.66 & 86.50 & 0.18 & 97.5 \\
    $\text{Ours}$                        & 0.67 & 86.49 & 0.18 & 93.1 \\
    \bottomrule
  \end{tabular}
\end{table}
\textbf{Length Reward Rule.}
Table \ref{tab:maincompare} shows that models trained with the length reward rule in RL generate more concise modeling code than those trained with SFT alone, while the cadrille baseline without this mechanism shows no improvement in conciseness. To exclude the influence of different RL algorithms, we trained a control model under the same GSPO framework without length rewards, setting reward weights as $\lambda_{\text{iou}}$=0.8 and $\lambda_{\text{val}}$=0.2. As shown in Table \ref{tab:length ablation}, this control model shows no improvement in ATL, while other metrics remain comparable to the main model. This indicates that our proposed length reward rule effectively improves the conciseness of generated modeling code with only a marginal impact on geometric accuracy and code validity. 

\begin{table}[t]
  \centering
  \caption{\textbf{Ablative evaluation of Networks for geometric prior images On
Deepcad dataset.} $\text{Ours}_{\text{3-vits}}$: extra ViT encoders;  $\text{Ours}_{\text{3-Projs}}$: extra projection layers; $\text{Ours}_{\text{crossAttn}}$: cross-attention fusion; Ours: shared encoder + concat.}
  \label{tab:Nets ablation}
  \begin{tabular}{lS[table-format=1.1]S[table-format=2.1]S[table-format=1.2]}
    \toprule
    Method & {IR (\%)$\downarrow$} & {IoU (\%)$\uparrow$} & {CD$\downarrow$} \\
    \midrule
    $\text{Ours}_{\text{3-vits}}$       & 1.82 & 72.97 & 0.43 \\
    $\text{Ours}_{\text{3-Projs}}$      & 1.86 & 73.51 & 0.43 \\
    $\text{Ours}_{\text{crossAttn}}$    & 1.93 & 72.56 & 0.45 \\
    $\text{Ours}$                       & 1.57 & 75.44 & 0.33 \\
    \bottomrule
  \end{tabular}
\end{table}

\textbf{Processing and Fusion networks for Geometric Priors}
Regarding the processing and fusion of geometric prior images, we experimented with three alternative network architectures under the same settings. In $\text{Ours}_{\text{3-vits}}$, depth and normal maps are processed by two additional ViT encoders initialized from the original visual encoder. In $\text{Ours}_{\text{3-Projs}}$, independent projection modules are introduced for depth and normal features. In $\text{Ours}_{\text{crossAttn}}$, geometric features are fused into RGB features via cross-attention before being fed into the LLM.

However, as shown in Table \ref{tab:Nets ablation}, all three structures result in performance degradation. We argue that the visual encoder of Qwen2VL has been sufficiently pretrained on large-scale image data and already generalizes well to various input types, including geometric images. Introducing insufficiently trained additional modules instead disrupts the original representation capacity of the model. Therefore, this paper adopts the simplest and most effective strategy---feeding all images (RGB, depth, normal) directly into the same visual encoder and projection layer, and concatenating them before inputting into the LLM.



\section{Conclusion}
We present a systematic post-training study for multimodal large language models (MLLMs) in the single-view CAD generation task, aiming to bridge the gap between visual perception and parametric modeling. Motivated by a theoretical analysis of how geometric priors can enhance the reconstruction of 3D structures from a single 2D observation, we are the first to inject both depth and surface normal cues into an MLLM-based CAD generation pipeline. This design explicitly aligns the model’s visual representations with underlying geometric consistency, enabling more accurate and physically plausible shape reasoning. To further improve the efficiency and interpretability of the generated modeling sequences, we introduce a group-wise length reward during reinforcement learning. Comprehensive experiments across multiple public benchmarks demonstrate that our method consistently outperforms existing approaches, establishing new state-of-the-art performance in single-view CAD generation. 
\bibliography{cvmbib}

\clearpage

\end{document}